\documentclass[conference,a4paper]{IEEEtran}
\IEEEoverridecommandlockouts

\usepackage[hidelinks]{hyperref}
\usepackage[cmex10]{amsmath}
\usepackage{amssymb,amsfonts}
\usepackage{dblfloatfix}

\usepackage[ruled,vlined]{algorithm2e}
\usepackage{graphicx}
\usepackage{graphics}
\usepackage{subcaption} 
\usepackage{multirow}
\graphicspath{{Figures/PDF/}{Figures/PNG/}}

\usepackage{booktabs}
\usepackage{siunitx}
\usepackage[numbers,compress]{natbib}
\usepackage{texnames}
\usepackage{bm,bbm}
\usepackage{orcidlink}
\usepackage{epsfig}
\usepackage{adjustbox}
\usepackage{setspace}

\begin{document}

\title{Task-Guided Multi-Annotation Triplet Learning for Remote Sensing Representations}

\author{
    \IEEEauthorblockN{Meilun Zhou\textsuperscript{1}\orcidlink{0000-0002-5891-3203}}
    \IEEEauthorblockA{\textsuperscript{1}Department of Electrical and Computer Engineering\\
    \textit{University of Florida}\\
    Gainesville, FL, USA, 32611\\
    zhou.m@ufl.edu}
    \and
    \IEEEauthorblockN{Alina Zare\textsuperscript{1,2}\orcidlink{0000-0002-4847-7604}}
    \IEEEauthorblockA{\textsuperscript{1}Department of Electrical and Computer Engineering\\
    \textsuperscript{2}Artificial Intelligence and Informatics Research Institute\\
    \textit{University of Florida}\\
    Gainesville, FL, USA, 32611\\
    azare@ufl.edu}
}


\maketitle
\begin{abstract}
Prior multi-task triplet loss methods relied on static weights to balance supervision between various types of annotation. However, static weighting requires tuning and does not account for how tasks interact when shaping a shared representation. To address this, the proposed task-guided multi-annotation triplet loss removes this dependency by selecting triplets through a mutual-information criteria that identifies triplets most informative across tasks. This strategy modifies which samples influence the representation rather than adjusting loss magnitudes. Experiments on an aerial wildlife dataset compare the proposed task-guided selection against several triplet loss setups for shaping a representation in an effective multi-task manner. The results show improved classification and regression performance and demonstrate that task-aware triplet selection produces a more effective shared representation for downstream tasks.
\end{abstract}


\section{Introduction}
Learning a shared representation for multi-task remote sensing requires combining heterogeneous supervisory signals into a single embedding space for downstream tasks \cite{houdre2025promm, li2025knowledge}. Class labels impose semantic structure, while box-level attributes provide geometric information about size, shape, and spatial layout. Each annotation captures distinct aspects of the data distribution but also introduce different constraints; semantic similarity may not align with geometric similarity, and geometric variation may not reflect semantic boundaries \cite{wang2016multi}. A shared representation must preserve both structures without distortion which require methods that integrate multiple annotation types to support tasks relying on semantic and geometric information simultaneously.

The multi-annotation discrete triplet loss \cite{zhou2025multi} (MATL) aligns samples using semantic and geometric supervision to learn embeddings that support both classification and spatial reasoning. MATL aggregates triplet losses computed from multiple annotation types by applying a fixed scalar weight to each task specific loss term. These weights control the relative influence of semantic labels versus geometric attributes during representation learning. Selecting appropriate values requires extensive manual tuning because the optimal balance depends on dataset characteristics, task difficulty, and annotation noise. A single global weighting scheme can overemphasize one form of supervision while suppressing others, which limits adaptability across tasks and datasets and motivates the need for more flexible task balancing strategies \cite{kendall2018multi, chen2018gradnorm}. Most multi-task balancing methods would address this by scaling loss magnitudes rather than modifying the representation itself and balancing gradients without shaping how different annotations influence relationships in the latent space.

The proposed method introduces a task-guided triplet selection mechanism that leverages mutual information between annotations to identify the most informative sample relationships. Task interactions determine which labels influence the representation and remove the need for a static weight. The selection mechanism emphasizes triplets that best capture agreement between semantic and geometric structure and adapt the embedding to these relationships. Unlike traditional balancing methods, which equalize gradients without affecting the latent space \cite{ahn2021use, li2022complex}, this approach modifies the triplet constraints themselves to produce embeddings that reflect interactions between label types rather than a weighted sum of independent losses.

Experiments train task-specific heads on embeddings from frozen Vision Transformer–based models and compare the proposed method against four setups: (1) base embeddings, (2) class-label triplet loss, (3) class-label triplet loss with hard triplet selection, and (4) multi-annotation triplet loss. Evaluations use a remote sensing wildlife dataset with multi-task annotations, including class labels and detailed box-level geometry to assess both classification and spatial reasoning performance.

\section{Proposed Methodology}
The proposed method builds on Vision Transformer–based model representations and multi-annotation triplet learning by introducing task guided triplet selection. Pretrained Vision Transformer–based models \cite{he2022masked, oquab2023dinov2, radford2021learning} are used as fixed feature extractors to generate initial embeddings. These embeddings serve as a common starting point for all compared methods, which are applied without updating the model encoders in order to isolate the effect of triplet selection on downstream performance. The approach begins with the multi-annotation triplet loss \cite{zhou2025multi}, which averaged separate triplet losses from class labels and box-derived labels. Instead of using a grid search to find the balance between these losses, we propose using mutual information \cite{strehl2002cluster} between class- and box-based similarities to identify the most informative sample triplets. This metric guides the selection of anchors, close-samples, and far-samples across annotations. The training objective uses the selected triplets in the multi-annotation triplet formulation. 

\subsection{Initial Embedding Extraction}
Embeddings are obtained from the aerial wildlife dataset using the pretrained Vision Transformer–based models. Each image patch is passed through the corresponding encoder to obtain a fixed-length feature vector. The model encoders remain frozen during all experiments to ensure that any changes in representation quality arise solely from the triplet selection process. This setup begins with state-of-the-art latent spaces and seeks to further isolate the effect of the proposed method on the learned latent space.

\subsection{Recap of Multi-Annotation Triplet Loss}
\label{sec:method_matl}

Let $\mathbf{x}_i \in \mathbb{R}^{d}$ denote the input image for sample $i$, $\mathbf{y}^{(i)}_{\mathrm{class}} \in \left\{1,\ldots,C\right\}$ the corresponding categorical class label for C classes, and $\mathbf{y}^{(i)}_{\mathrm{box}} \in \mathbb{R}^{4}$ the associated bounding box annotation. Let
\[
\mathcal{D}
=
\left\{
\left(\mathbf{x}_i,\mathbf{y}^{(i)}_{\mathrm{class}},\mathbf{y}^{(i)}_{\mathrm{box}}\right)
\right\}_{i=1}^{N}
\]
denote a dataset of $N$ samples.

Let $f:\mathbb{R}^{d}\rightarrow\mathbb{R}^{k}$ denote an embedding network that maps each input $\mathbf{x}_i$ to a $k$ dimensional latent representation.  
Let $d:\mathbb{R}^{k}\times\mathbb{R}^{k}\rightarrow\mathbb{R}$ denote a distance metric and let $\alpha>0$ denote a margin parameter. A triplet consists of three input samples
\[
\left(\mathbf{x}_a,\mathbf{x}_p,\mathbf{x}_n\right),
\]
where $\mathbf{x}_a$ is an anchor sample, $\mathbf{x}_p$ is selected as a sample with the same label as the anchor, and $\mathbf{x}_n$ is selected as a sample with a different label to the anchor.  
The standard triplet loss is defined as
\begin{equation}
\label{equation:basetriplet}
\begin{aligned}
\mathcal{L}_{\mathrm{triplet}}(\mathbf{x}_a,\mathbf{x}_p,\mathbf{x}_n)
=
\max\!\Big(
& d\!\big(f(\mathbf{x}_a),f(\mathbf{x}_p)\big)
\\
& {}-
d\!\big(f(\mathbf{x}_a),f(\mathbf{x}_n)\big)
+
\alpha,
\,0
\Big).
\end{aligned}
\end{equation}

\noindent
Multi-annotation triplet loss extends this formulation by constructing triplets using multiple supervision sources.

A class-guided triplet loss is defined as
\begin{equation}
\label{equation:classtriplet_concise}
\mathcal{L}_{\mathrm{class}}(\mathbf{x}_a,\mathbf{x}_p,\mathbf{x}_n)
=
\mathcal{L}_{\mathrm{triplet}}\!\left(
\mathbf{x}_a,\mathbf{x}_p,\mathbf{x}_n
\,\middle|\,
\mathbf{y}_{\mathrm{class}}
\right),
\end{equation}
where triplet selection is determined by semantic class membership. From the continuous box annotations $\mathbf{y}_{\mathrm{box}}$, a discrete geometric label
\[
\tilde{\mathbf{y}}^{(i)}_{\mathrm{box}} \in \{1,\dots,B\}
\] is constructed using a predefined discretization scheme over the box feature space. A box-guided triplet loss is then defined as
\begin{equation}
\label{equation:boxtriplet_concise}
\mathcal{L}_{\mathrm{box}}(\mathbf{x}_a,\mathbf{x}_p,\mathbf{x}_n)
=
\mathcal{L}_{\mathrm{triplet}}\!\left(
\mathbf{x}_a,\mathbf{x}_p,\mathbf{x}_n
\,\middle|\,
\tilde{\mathbf{y}}_{\mathrm{box}}
\right),
\end{equation}
where triplet selection is determined by discrete geometric label agreement.

The combined multi-annotation triplet loss is given by
\begin{equation}
\label{equation:MATL_concise}
\begin{aligned}
\mathcal{L}_{\mathrm{MATL}}(\mathbf{x}_a,\mathbf{x}_p,\mathbf{x}_n)
=
& (1-\lambda)\,
\mathcal{L}_{\mathrm{class}}(\mathbf{x}_a,\mathbf{x}_p,\mathbf{x}_n)
\\
& {}+
\lambda\,
\mathcal{L}_{\mathrm{box}}(\mathbf{x}_a,\mathbf{x}_p,\mathbf{x}_n),
\end{aligned}
\end{equation}
where $\lambda\in[0,1]$ controls the relative contribution of semantic and geometric supervision and is set by the user. $\lambda$ is set to 0.5 in this study. 

\subsection{Mutual Information Triplet Selection and Sample Masking}
\label{subsec:mi_guided_selection}

Let $\mathbf{Y}_{\mathrm{box}}\in\mathbb{R}^{N\times F}$ denote the min-max normalized geometric box feature matrix, where $F=4$ and where $[\mathbf{Y}_{\mathrm{box}}]_{if}$ denotes the value of box feature $f$ for sample $i$.  A per sample relevance score $m_i$ is defined as
\begin{equation}
\label{eq:per_sample_metric}
m_i
=
\sum_{f=1}^{F}
[\mathbf{Y}_{\mathrm{box}}]_{if}\,
\mathrm{MI}\!\big(
\mathbf{y}_{\mathrm{box},f},
\mathbf{y}_{\mathrm{class}}
\big),
\end{equation}
where $\mathrm{MI}(\cdot,\cdot)$ denotes the mutual information between a continuous feature vector and the categorical class label vector.

For each class $c\in\{1,\dots,C\}$ define the index set

\begin{equation}
\mathcal{I}_c
=
\{\, i \mid y^{(i)}_{\mathrm{class}} = c \,\}.
\end{equation}

\noindent Let $p_{\mathrm{top}}\in(0,1]$ denote the top fraction.  
The number of top ranked samples is

\begin{equation}
n_{\mathrm{top},c}
=
\max\!\Big(
1,\,
\lfloor p_{\mathrm{top}}\,|\mathcal{I}_c| \rfloor
\Big).
\end{equation}

\noindent Then we define the set of top ranked indices as

\begin{equation}
\mathcal{T}^{\mathrm{top}}_c
=
\operatorname{TopIndices}
\big(
\{m_i\}_{i\in\mathcal{I}_c},
\,n_{\mathrm{top},c}
\big),
\end{equation}

\noindent and let the remaining indices be

\begin{equation}
\mathcal{R}_c
=
\mathcal{I}_c \setminus \mathcal{T}^{\mathrm{top}}_c.
\end{equation}

Let $p_{\mathrm{rand}}\in[0,1]$ denote the random fraction.  
The number of random samples is

\begin{equation}
n_{\mathrm{rand},c}
=
\max\!\Big(
0,\,
\lfloor p_{\mathrm{rand}}\,|\mathcal{I}_c| \rfloor
\Big).
\end{equation}

\noindent A random subset is sampled without replacement

\begin{equation}
\mathcal{T}^{\mathrm{rand}}_c
\subseteq
\mathcal{R}_c,
\qquad
|\mathcal{T}^{\mathrm{rand}}_c|
=
\min\!\big(
n_{\mathrm{rand},c},
|\mathcal{R}_c|
\big),
\end{equation}

\noindent and the final selected indices for class $c$ are

\begin{equation}
\mathcal{T}_c
=
\mathcal{T}^{\mathrm{top}}_c
\cup
\mathcal{T}^{\mathrm{rand}}_c.
\end{equation}

\noindent Finally aggregating across all classes yields a binary mask $\mathcal{M}\in\{0,1\}^N$ defined by

\begin{equation}
\mathcal{M}_i
=
\begin{cases}
1 & \text{if } i\in\bigcup_{c=1}^{C}\mathcal{T}_c,\\
0 & \text{otherwise.}
\end{cases}
\end{equation}

Let $\mathcal{T}_{\mathrm{selected}}$ denote the set of triplets $(a,p,n)$ constructed using anchor and positive indices with $\mathcal{M}_i=1$ and negatives sampled according to the multi-annotation strategy.  
The task guided multi-annotation triplet loss is

\begin{equation}
\label{eq:MATL_TG_final}
\mathcal{L}_{\mathrm{MATL\text{-}TG}}=\sum_{\mathcal{T}_{\mathrm{selected}}}\mathcal{L}_{\mathrm{MATL}}(a,p,n),
\end{equation}

\noindent where $\mathcal{L}_{\mathrm{MATL}}$ is defined in Equation~\ref{equation:MATL_concise}.

\begin{figure}[!htb]
    \centering

    \includegraphics[width=0.95\linewidth]{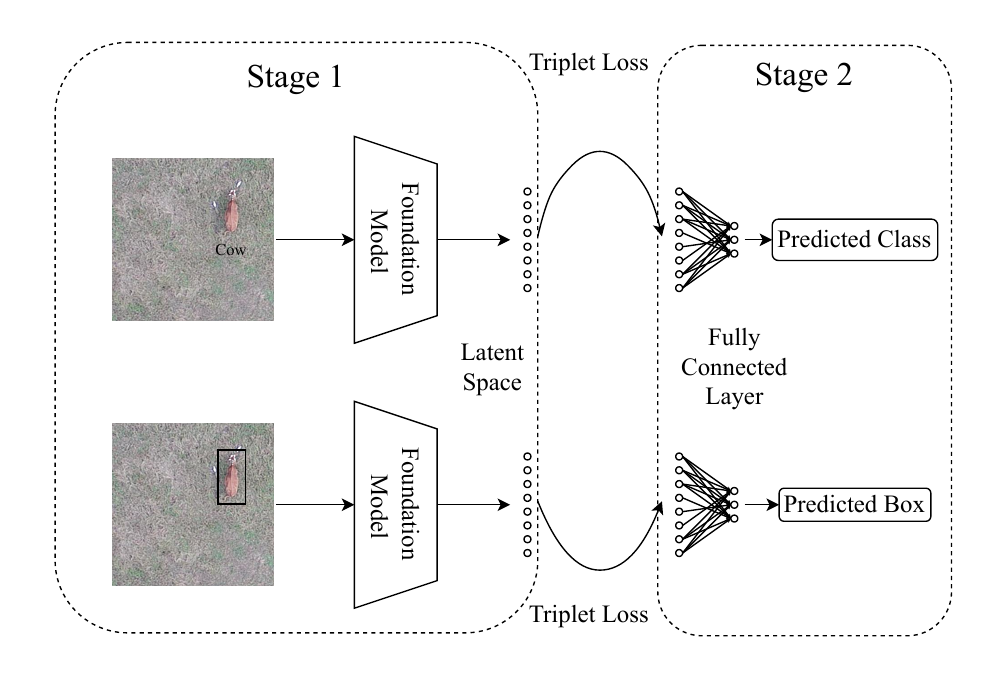}
    \caption{Two-stage architecture. Stage one extracts frozen Vision Transformer–based model embeddings and applies different representation learning losses to shape the latent space. Stage two trains linear task heads on each latent space to measure the impact on downstream performance.}
    \label{fig:awir_arch}
\end{figure}

\section{Experimental Results}
This work uses the Animal Wildlife Image Repository (AWIR) dataset from the study \cite{zhou2025multi}. The original data preparation pipeline included image tiling, target isolation, annotation processing, and feature extraction. The same methodology is applied in this paper to ensure consistency across experiments. The area and symmetric squareness features are derived from the box annotations for this study. This dataset supports the experiments in this paper because the annotations include semantic information and geometric structure. 

All models were trained using NVIDIA B200 GPU with $12$ CPUs and $60$ GB of memory using Tensorflow. Code for these experiments are provided at \url{https://github.com/GatorSense/Task-Guided-MATL}.

The task guided framework further benefits from quantifying statistical dependence between heterogeneous sources of supervision. 
Mutual information \cite{strehl2002cluster} provides a general measure of dependency between two random variables without assuming linearity or a specific parametric form. 
For two arbitrary random variables $U$ and $V$, mutual information is defined as

\begin{equation}
\label{eq:general_mi}
\mathrm{MI}(U,V)
=
\sum_{u\in\mathcal{U}}
\sum_{v\in\mathcal{V}}
p(u,v)
\log
\frac{
p(u,v)
}{
p(u)\,p(v)
}.
\end{equation}

This general formulation is used in this work by setting $U$ to individual geometric feature dimensions derived from bounding boxes and $V$ to semantic supervision signals from class labels. 
The resulting mutual information values quantify the degree to which geometric structure aligns with semantic variation. 
These statistics are subsequently used to construct the per sample relevance scores defined in Equation~\ref{eq:per_sample_metric}, which guide triplet selection toward samples that encode strong cross task relationships. 
Table~\ref{table:mi_stats}. summarizes the distribution of mutual information values across categories.

\begin{table}[!htb]
\caption{Sample-wise mutual information for AWIR dataset. Values summarize how strongly the area and symmetric squareness attributes align with species-level semantic categories.}
\begin{center}
\begin{tabular}{|c | c | c | c|} 

 \hline
 \textbf{Category} & \textbf{Cow} & \textbf{Deer} & \textbf{Horse}\\ [0.5ex] 
 \hline\hline
\textbf{Sample Count} & 116 & 51 & 73\\ 
 \hline
\textbf{Mean} & 0.180 & 0.115 & 0.374\\ 
 \hline
\textbf{Std. Dev.} & 0.096 & 0.039 & 0.194\\ 
 \hline
\textbf{Min} & 0.028  & 0.046  & 0.1105\\
 \hline
\textbf{Max} & 0.630 & 0.214 & 0.744 \\
 \hline
 
\end{tabular}
\end{center}
\label{table:mi_stats}
\end{table}

Five setups are evaluated to isolate the effect of task-guided triplet sampling. The first setup uses the frozen Vision Transformer–based encoder embeddings to establish a baseline performance level from supervised training of the task-heads alone. The second setup applies triplet learning with class labels and measures how discrete class label supervision shapes the latent space. The third setup uses the same class label based triplet learning but with an additional hard selection requirement \cite{xuan2020improved}. The fourth setup uses the multi-annotation triplet loss and combines class and box annotation with a fixed $0.5$ per weight scheme to measure how multiple supervision types influence the representation when the loss design remains static. The final setup applies the proposed task-guided strategy. This approach keeps the fixed loss weighting and adds the sampling process driven by the mutual information relationship between semantic and geometric annotations. The experiment architecture is outlined in Fig. \ref{fig:awir_arch}.

\begin{table*}[htpb]
\caption{This table reports classification accuracy and box feature regression $R^2$ for five training setups. The setups include Without Triplet Loss (WTL), Class Label Triplet Loss (CLTL), Class Label Triplet Loss with hard triplet selection (CLTL hard), Multi-Annotation Triplet Loss (MATL) with weight 0.5, and TG-MATL as the proposed method. Average and standard deviation values of eight runs of each setup are shown across three Vision Transformer–based models embeddings: DINOv2 and CLIP as well as MAE.}

\setlength{\tabcolsep}{3pt} 
\begin{center}
\adjustbox{max width=\textwidth}{ 
\begin{tabular}{||c | c | c | c | c | c | c | c | c | c | c||}
 \hline
 \multirow{3}{*}{\textbf{Model}}  & \multicolumn{5}{c}{\textbf{Classification (Overall Accuracy)}} & \multicolumn{5}{|c||}{\textbf{Box Feature Regression ($R^2$)}} \\ [0.5ex] 
 \cline{2-11}
  & \multirow{2}{*}{WTL}  & \multirow{2}{*}{CLTL}  & \multirow{2}{*}{CLTL}   & \multirow{2}{*}{MATL} & \multirow{2}{*}{\textbf{TG-MATL}}     & \multirow{2}{*}{WTL}  & \multirow{2}{*}{CLTL}  & \multirow{2}{*}{CLTL}   & \multirow{2}{*}{MATL} & \multirow{2}{*}{\textbf{TG-MATL}}   \\ [0.5ex] 
  &  &  & \textit{(Hard)} & \textit{(0.5)} &  & &  &  \textit{(Hard)} & \textit{(0.5)} &\\ [0.5ex] 

 \hline\hline

 DINOv2 & $96.28 \pm 0.76$ & $96.35 \pm 0.67$ & $\mathbf{96.87 \pm 0.27}$ & $96.43 \pm 0.84$ &  $96.80 \pm 0.78$  & $0.63 \pm 0.03$ & $0.58 \pm 0.04$ & $0.60 \pm 0.04$ & $0.61 \pm 0.02$ & $\mathbf{0.64 \pm 0.01}$  \\
\hline

 MAE & $\mathbf{82.22 \pm 1.92}$  & $78.79 \pm 1.93$ & $79.38 \pm 1.34$  & $77.60\pm 1.93$ & $79.99 \pm 2.79$  & $\mathbf{0.44 \pm 0.04}$ & $0.25 \pm 0.02$ & $0.24 \pm 0.05$ & $0.29 \pm 0.02$ & $0.32 \pm 0.03$  \\ 
\hline

 CLIP & $89.58 \pm 0.84$  & $94.12 \pm 2.23$ & $\mathbf{95.08 \pm 1.09}$  & $93.53 \pm 2.81$ & $93.15 \pm 2.62$  & $\mathbf{0.60 \pm 0.01}$ & $0.45 \pm 0.04$ & $0.42 \pm 0.02$ & $0.54 \pm 0.04$ & $0.58 \pm 0.03$  \\ 
 \hline

\end{tabular}
}
\end{center}
\label{table:results}
\end{table*}

This experiment evaluates classification and box regression performance, with results summarized in Table~\ref{table:results}. For MATL, the best-performing top and random-percent sampling ratio is reported. Task-guided MATL (TG-MATL) improves classification performance over standard MATL by incorporating task-relevant sampling, which better aligns the representation with classification objectives. However, TG-MATL does not surpass hard-sampled CLTL, which achieves the highest classification accuracy due to the strong bias toward separating class distinctions. This gain in classification performance comes at the expense of regression, as the hard sampling strategy reduces consistency with continuous geometric relationships. 

For box regression, TG-MATL produces the highest $R^2$ value with DINOv2. In many cases, the absolute $R^2$ under TG-MATL remains slightly lower than the no-triplet baseline. Despite this trend, TG-MATL consistently improves box regression $R^2$ compared to both CLTL and standard MATL across all embedding models. These results suggest that task-guided sampling strengthens geometric reasoning while minimizing loss of semantic structure and reduces the trade-offs introduced by fixed triplet definitions formed by discrete labels.

\subsection{Top and Random Percent Experiment}
An additional study evaluates the sensitivity of the method to the sampling hyperparameters by examining downstream task performance under different values for the top-percent and random-percent ratios. This experiment is performed for the CLIP embeddings.  

\begin{figure}[htpb]
    \centering
    \includegraphics[width=1.0\linewidth]{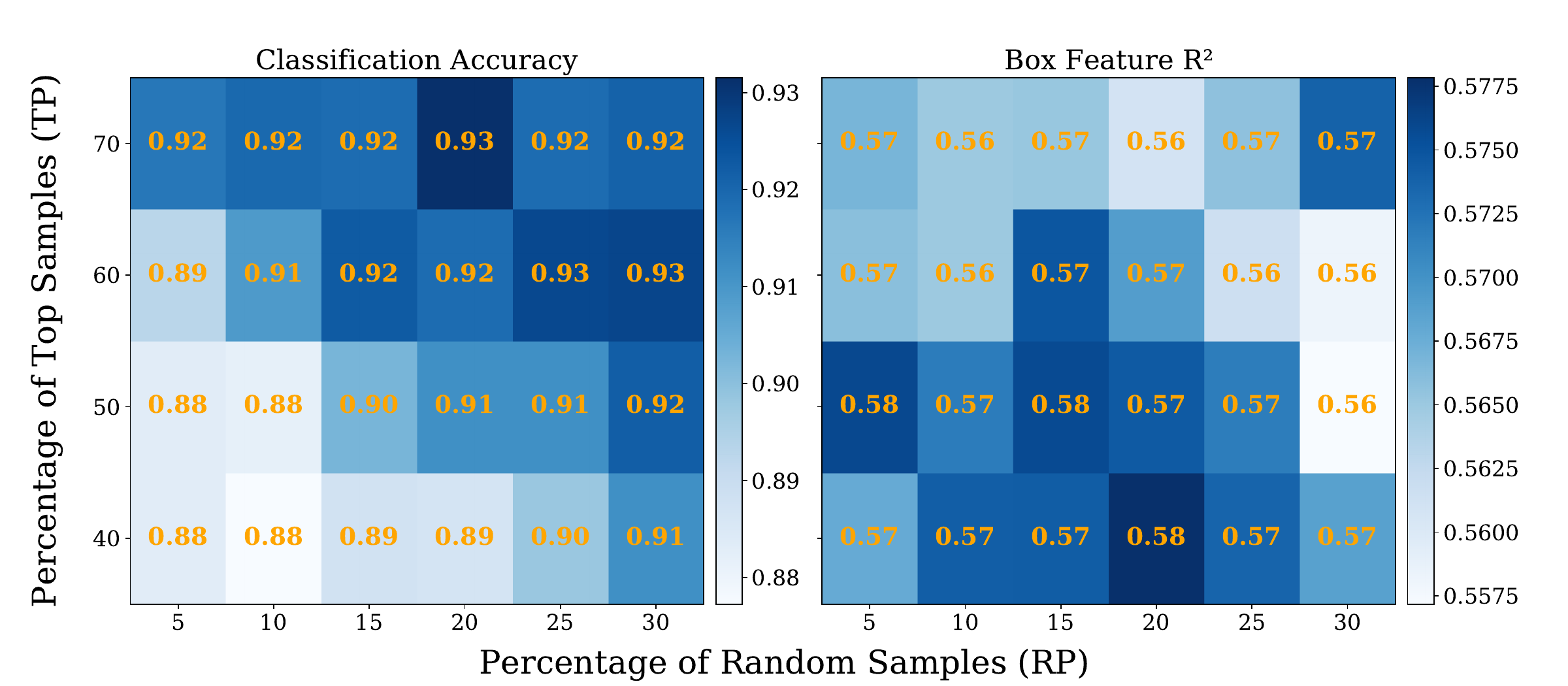}
    \caption{
        The figure presents two heatmaps that summarize how the percentage of top samples and the percentage of random samples impact downstream task performance for TG-MATL on CLIP embeddings. All reported values are the average over eight runs of each experiment.
    }
    \label{fig:tp_rp_heatmaps}
\end{figure}

The results reveal a consistent trend for classification, where increasing the proportion of random samples improves accuracy. Broader random sampling increases intra class diversity, which strengthens semantic separation for an easier discriminative task. In contrast, increasing the random percentage yields limited improvement and can degrade box feature $R^2$ which suggests greater sensitivity to weak or noisy relationships. The box task may be more difficult and the dataset may have fewer highly informative samples, so adding more random samples dilutes supervision rather than strengthening. Performance for box regression is highest when the top percent is small, since stricter selection emphasizes reliable geometric relationships.

\section{Discussion}
Mutual information \cite{strehl2002cluster} measures how strongly different annotation types align, where high values indicate shared structure and low values indicate weak relationships. Task-guided selection uses these relationships to prioritize informative triplets across annotations rather than optimizing each task independently. Controlling triplet selection at this stage directly shapes sample relationships in the representation, whereas weighting losses afterward cannot correct uninformative or misleading comparisons once they have already influenced the embedding.

CLIP, DINOv2, and MAE are all Vision Transformer-based models but differ in training objectives and data sources. CLIP uses contrastive learning on paired image and text data, which aligns visual and semantic representations across modalities and provides supervision through text-derived labels. DINOv2 relies on self-distillation on large curated image datasets, which captures strong visual structure without explicit labels. MAE uses a reconstruction objective that masks portions of the input image and predicts missing content, which emphasizes structural information. 

CLIP and DINOv2 both provide strong starting representations for MATL, since CLIP benefits from label-informed alignment and DINOv2 captures rich visual structure with high-capacity embeddings. Among these approaches, DINOv2 often produces the strongest visual representation due to the training strategy and higher-dimensional embeddings, which makes DINOv2 particularly effective as a foundation for structuring the latent space.

\section{Conclusion}
This work replaces static weighted losses with dynamic task-aware triplet selection. The method identifies the sample relationships that hold value for multiple annotations and shapes the latent space through those relationships. 

Future research can extend this style of representation learning towards a more continuous and smooth latent space. This direction would allow the model to construct triplets using real-valued geometric attributes without discretization.  Additionally, this work could be applied to multi-modal settings \cite{dutt2022shared} involving heterogeneous sources of data that would benefit from a shared representation.

\small
\bibliographystyle{IEEEtranN}
\bibliography{references}

\end{document}